\documentclass{article}
\usepackage{placeins}
\usepackage{PRIMEarxiv}
\usepackage{float}
\usepackage[utf8]{inputenc} 
\usepackage[T1]{fontenc}    
\usepackage{hyperref}       
\usepackage{url}            
\usepackage{booktabs}       
\usepackage{amsfonts}       
\usepackage{nicefrac}       
\usepackage{microtype}      
\usepackage{lipsum}
\usepackage{fancyhdr}       
\usepackage{graphicx}       
\graphicspath{{media/}}     
\usepackage{multirow}
\usepackage{booktabs}
\title{ERNIE-UniX$^2$: A Unified Cross-lingual Cross-modal Framework for Understanding and Generation}

\author{
Bin Shan \footnotemark[1]
\hspace{1cm}Yaqian Han \thanks{indicates equal contribution.}
\hspace{1cm}Weichong Yin
\hspace{1cm}Shuohuan Wang
\hspace{1cm}Yu Sun\\
\\
\textbf{\hspace{1cm}Hao Tian
\hspace{1cm}Hua Wu
\hspace{1cm}Haifeng Wang}
\\\\\texttt{Baidu Inc., China}
\\\\
\texttt{\{shanbin01,hanyaqian,yinweichong,wangshuohuan\}@baidu.com}
}

\begin{document}
\maketitle

\begin{abstract}
Recent cross-lingual cross-modal works attempt to extend Vision-Language Pre-training (VLP) models to non-English inputs and achieve impressive performance. However, these models focus only on understanding tasks utilizing encoder-only architecture. In this paper, we propose ERNIE-UniX$^2$, a unified cross-lingual cross-modal pre-training framework for both generation and understanding tasks. ERNIE-UniX$^2$ integrates multiple pre-training paradigms (e.g., contrastive learning and language modeling) based on encoder-decoder architecture and attempts to learn a better joint representation across languages and modalities. Furthermore, ERNIE-UniX$^2$ can be seamlessly fine-tuned for varieties of generation and understanding downstream tasks. Pre-trained on both multilingual text-only and image-text datasets, ERNIE-UniX$^2$ achieves SOTA results on various cross-lingual cross-modal generation and understanding tasks such as multimodal machine translation and  multilingual visual question answering.  
\end{abstract}

\section{Introduction}
Pre-trained cross-modal models \cite{simvlm2021,dou2022an,chen2020uniter,Jia2021ScalingUV,vilbert2019,vlt52021,albef2021} have achieved impressive results across various vision-language understanding and generation tasks (e.g., image-text retrieval, visual question answering, image captioning). While most works are based on datasets in English,  their models can not be easily used in cross-modal applications of non-English languages. While we are living in a multilingual world, it is worth exploring to build a unified model across languages and modalities. 

To tackle cross-modal tasks across languages, researchers have proposed cross-lingual cross-modal frameworks and achieved promising results, such as M$^{3}$P \cite{m3p2020}, and UC$^2$ \cite{UC22021}.
However, attempting to learn cross-lingual cross-modal representation through pre-training with bidirectional attention in encoder-only architecture, these frameworks can not easily be applied for generation tasks, such as multimodal machine translation. 
\begin{figure}
\centering\includegraphics [scale=0.5]{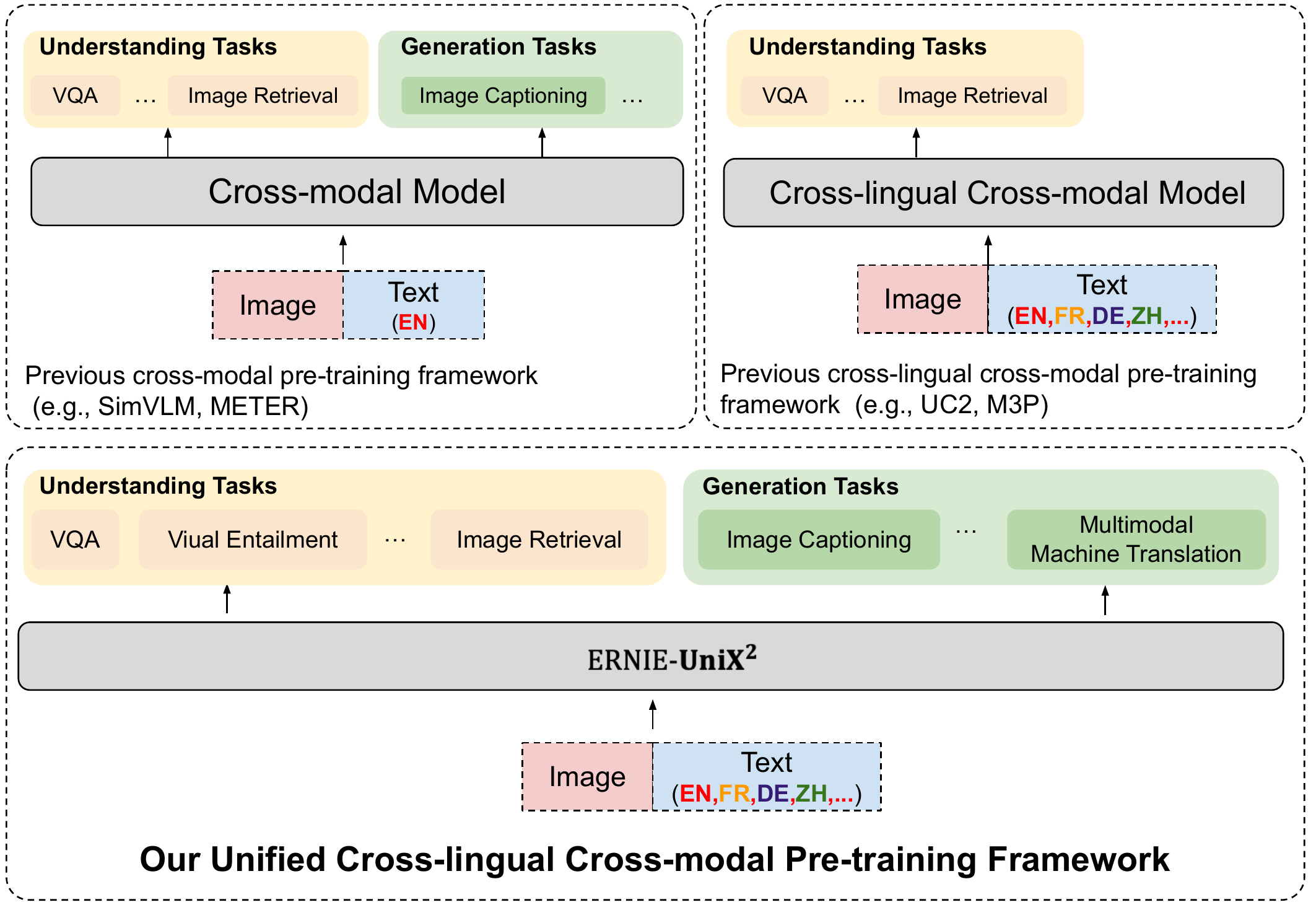}
\caption{While existing unified cross-modal pre-trained models are mainly trained on English datasets and recent cross-lingual cross-modal pre-trained models can only be applied to understanding tasks, our proposed \textbf{ERNIE-UniX$^2$} is a unified pre-training framework, capable of tackling both understanding and generation tasks of cross-lingual cross-modal scenarios.  }
\label{fig:figure1_cap}
\end{figure} 

In this paper, we propose a unified cross-lingual cross-modal framework ERNIE-UniX$^{2}$ integrating multiple generations and understanding pre-training paradigms,  which can be seamlessly fine-tuned on downstream tasks. Figure \ref{fig:figure1_cap} illustrates the main difference between ERNIE-UniX$^{2}$ and previous cross-modal and cross-lingual cross-modal models.
ERNIE-UniX$^{2}$ is built based on the encoder-decoder architecture, where the encoder fuse cross-lingual cross-modal feature with a token level interaction between aligned visual and language representation extracting two separated image/text encoder.
To pre-train ERNIE-UniX$^{2}$, we integrate multiple pre-training tasks to learn a joint cross-lingual cross-modal representation, focusing on both the capabilities for understanding and generation. 
For understanding pre-training tasks, we perform contrastive learning with dual-encoder architecture on bilingual texts and image-text pairs to learn global-level alignments across different languages and modalities, and feed the features to a cross-lingual cross-modal encoder to further learn the fine-grained alignments using two widely-used pre-training tasks in VLP (i.e., masking language modeling and image-text matching).
To enhance the capabilities of our framework on generation tasks, we extend Prefix Language Modelling (PrefixLM) \cite{simvlm2021} to the multilingual scenario. While the PrefixLM only learns on monolingual inputs, we further adopt bilingual generation pre-training tasks to learn the joint alignment between bilingual texts and visual context. 
Besides, our framework adopts a more efficient image backbone ViT \cite{vit2020} with patch-level visual features instead of region-level visual features by a heavy object detector in previous works \cite{m3p2020,UC22021}

We utilize text-only datasets (CC100 \cite{xlmr2019} and OPUS \cite{opus-2012}) and image-text datasets (CC3M \cite{coco3m}, CC12M \cite{cc12m} and WIT \cite{wit2021}) to train ERNIE-UniX$^2$, in which we use translated CC3M and CC12M following \cite{UC22021}\footnote{We directly use translated CC3M provided in \cite{UC22021} and translate captions of CC12M by public translate API https://fanyi-api. baidu.com/}. Transferred to understanding and generation downstream tasks, ERNIE-UniX$^{2}$ achieves SOTA results for cross-lingual image-text retrieval, multimodal machine translation on Multi30K \cite{elliott-EtAl:2017:WMT,barrault2018findings}, cross-lingual visual language reasoning on XVNLI \cite{xvnli-lingual}. Moreover, compared with previous SOTA VLP models pre-trained on large-scale datasets in well-resourced languages, ERNIE-UniX$^2$ presents competitive results on image captioning on MSCOCO \cite{mscoco} and COCO-CN \cite{cococn}.

Overall, our contributions fall into two parts: 
\begin{enumerate}
\item We propose ERNIE-UniX$^{2}$, the first unified cross-lingual cross-modal framework which can be applied to both generation and understanding tasks.
\item ERNIE-UniX$^{2}$ achieves SOTA results on both cross-lingual cross-modal understanding tasks (e.g., XVNLI) and generation tasks (Multimodal Machine Translation), and presents competitive results on image captioning in well-resourced languages. 
\end{enumerate}

\section{Related Work}
\paragraph{Cross-modal Pre-trained Models}
In the past three years, numerous works \cite{vilbert2019,chen2020uniter,simvlm2021,Li2020OscarOA,Kim2021ViLTVT,Jia2021ScalingUV,dou2022an} in Vision-Language Pre-training have shown their effectiveness in learning generic cross-modal representation and significantly improved the performance of various downstream cross-modal tasks. These works roughly fall into two categories. 

The first category exploits cross-modal attention to learn fine-grained alignments between visual and textual tokens and presents superior performances on cross-modal tasks (especially for fine-grained vision-language tasks). However, most early works utilize region-level visual features with pre-trained detectors, which requires complicated pre-processing in training and inference. Motivated by an advanced image backbone ViT \cite{vit2020}, recent works \cite{Kim2021ViLTVT,simvlm2021,dou2022an} propose more efficient pre-training frameworks directly extracting visual features via a simple linear projection of patches. Among these works, SimVLM \cite{simvlm2021} proposes a simple yet effective pre-training task PrefixLM and achieves impressive results on cross-modal tasks with an encoder-decoder architecture, exploiting bi-directional and left-to-right attention across languages and modalities simultaneously. Another category of work (e.g., ALIGN \cite{Jia2021ScalingUV} and CLIP \cite{Radford2021LearningTV}) attempts to align vision and language using contrastive learning with parallel encoders (dual-encoder) and obtain robust vision and vision-language representation only leveraging web-crawled image-text datasets. However, these works utilize shallow cross-modal interactions, resulting in low performance on fine-grained vision-language tasks (e.g., VQA) requiring deeper multimodal understanding. To address these limitations, recent works \cite{Yuan2021FlorenceAN,Singh2021FLAVAAF,albef2021} attempt to integrate contrastive learning and cross-modal attention with fine-grained interactions into one pre-training framework. In our framework, we integrate dual-encoder with contrastive learning into encoder-decoder architecture to be flexibly applied to various downstream tasks and focus on improving the performance of both cross-lingual cross-modal generation and understanding tasks.

\paragraph{Cross-lingual Cross-modal Pre-trained Models}
Cross-lingual cross-modal pre-training aims to generalize such success achieved by VLP to a wide range of non-English inputs. The pioneer work M$^3$P \cite{m3p2020} proposes cross-modal Code-switched Training (MCT) to learn a joint cross-lingual cross-modal representation leveraging English-only image-text pairs. Since MCT aligns images and multilingual texts with English as a pivot, UC$^2$ \cite{UC22021} extended to multiple pivots (vision and languages) for enhancing the alignment by proposing Visual Translation Language Modeling (VTLM) and translating existing English-only image-text pairs into other languages. On the other hand, motivated by recent works on cross-modal retrieval using contrastive learning, MURAL \cite{mural2021} achieves significant improvements on cross-lingual image-text retrieval through pre-training on large-scale web-crawled multilingual image-text pairs with the dual-encoder architecture. While these works focus on understanding tasks, we attempt to tackle understanding and generation tasks with a unified pre-training framework across languages and modalities. 
\section{Method}
In this section, we present a cross-lingual cross-modal framework ERNIE-UniX$^{2}$ unifying generation and understanding tasks. We present the overview of our frameworks in Section \ref{Framework}, then describe the pre-training tasks in Section \ref{Objectives}.
\begin{figure}[h]
\centering\includegraphics [scale=0.56] {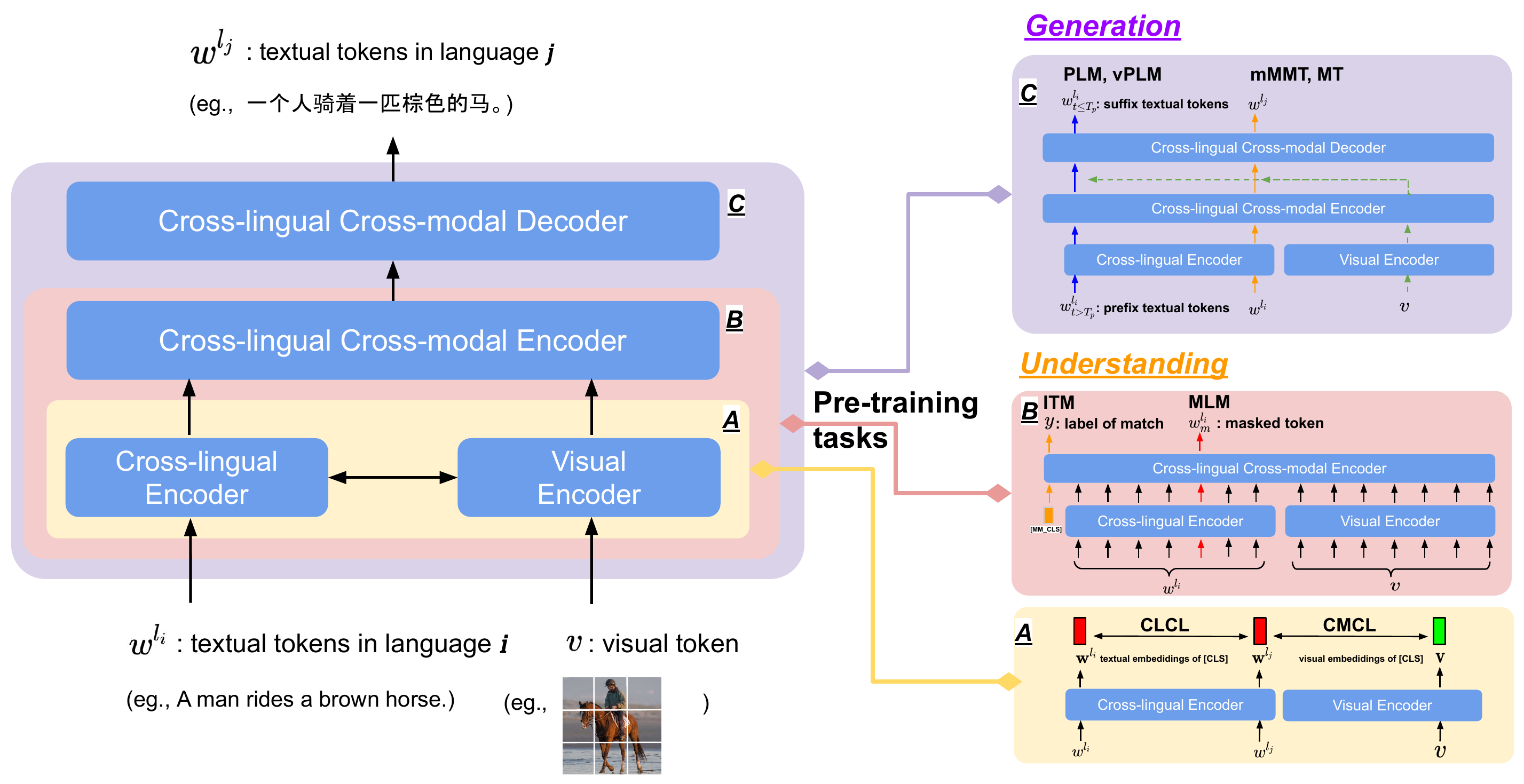}
\caption{The overview of our proposed \textbf{ERNIE-UniX$^{2}$} framework. \textbf{ERNIE-UniX$^{2}$} integrates three pre-training paradigms on different components. 1) Pre-training on Component. \textbf{\textit{A}} (dual-encoder architecture) with Cross-Lingual Contrative Learning (\textbf{CLCL}) for bilingual parallel texts and Cross-modal Contrastive Learning (\textbf{CMCL}) for text-image pairs. 2) Pre-training on Component. \textbf{\textit{B}} ( dual-encoder and encoder-decoder architecture) with Prefix Language Modeling (\textbf{PLM})  and Machine translation (\textbf{MT}) on cross-lingual texts, visual Prefix Language Modeling (\textbf{vPLM})  and Multimodal Machine Translation with masked textual tokens (\textbf{mMMT}) on cross-lingual image-text pairs. 3) Pre-training on Component. \textbf{\textit{C}} (encoder-only architecture) with Masked Language Modelling (\textbf{MLM}) and Image Text Matching (\textbf{ITM}).$T_{p}$ denotes the length of prefix tokens.}
\label{fig:framework}
\end{figure} 
\subsection{Framework Overview}
\label{Framework}
As illustrated in Figure \ref{fig:framework}, we build ERNIE-UniX$^{2}$ based on encoder-decoder architecture and incorporate three pre-training paradigms into a unified framework, which could seamlessly apply to varieties of understanding and generation downstream tasks with a unified framework across different languages.

For understanding tasks, we employ Cross-lingual Contrastive learning (CLCL) and  Cross-Modal Contrastive Learning (CMCL) with the dual-encoder for parallel bilingual text pairs and image-text pairs, respectively, to learn the coarse-grained alignment with shallow cross-modal interactions. Moreover, we exploit two widely-used Image-Text Matching (ITM) and Masked Language Modeling (MLM) to further learn a fine-grained vision-language alignment with the encoder. For generation tasks, we employ four pre-training tasks leveraging encoder-decoder, i.e.,  Prefix Language Modeling (PLM), visual Prefix Language Modeling (vPLM), Machine Translation (MT) and masked Multimodal Machine Translation (mMMT). 
vPLM and PLM attempt to enhance the generation capabilities of single languages with and without visual condition, and mMMT and MT aims to extend it to bilingual texts/text-image pairs.

We note that, we adopt an end-to-end vision Transformer as our visual encoder to extract patch-level features from images instead of relying on region-level visual features with a pre-trained detector (e.g., BUTD \cite{anderson2018bottom}) used in previous works M$^3$P \cite{m3p2020} and UC$^2$ \cite{UC22021}. 
\subsection{Pre-training Tasks}
\label{Objectives}
ERNIE-UniX$^2$ is trained on all pre-training tasks jointly with cross-lingual cross-modal mixed data stream sharing the same parameters. We construct data streams $D$ consisting of two main parts: text-only data stream $D^{[\mathrm{L}]}$ and image-text data stream $D^{\mathrm{VL}}$. In each iteration, a specified task is selected to calculate its gradients with equal probabilities, and then we merge all the gradients for all tasks to update the parameters.  
\subsubsection{Cross-lingual Cross-modal Contrastive Learning}
\label{Understanding}
We adopt Cross-modal Contrastive Learning (CMCL) on cross-lingual text-image pairs and Cross-lingual Contrastive Learning (CLCL) on cross-lingual text pairs to learn robust representations across languages and modalities. Overall, contrastive learning attempts to pull the matched image-text pairs together and push the others apart in a batch. Specifically, CMCL learns a cross-modal alignment with loss as
\begin{equation}
\mathcal{L}_{CMCL}=\mathbb{E}_{(w^{l_i},v)\sim D^{[\mathrm{VL}]}}-\frac{1}{N}(\underbrace{ \sum_{a}^{N}\log\frac{{\exp{(\mathbf{w}_{a}^{l_i}}^\top}\mathbf{v}_{a}/\tau)}{\sum_{b}^{N}\exp{({\mathbf{w}_{a}^{l_i}}^{\top}\mathbf{v}_b/\tau)}}}_{\mathrm{text-to-image}}+\underbrace{\sum_{a}^{N}\log\frac{{\exp{(\mathbf{v}_{a}^\top}}\mathbf{w}^{l_i}_{a}/\tau)}{\sum_{b}^{N}\exp{({\mathbf{v}_a^\top}\mathbf{w}^{l_i}_b/\tau)}}}_{\mathrm{image-to-text}})  
\end{equation}
where $\mathbf{w}_{a}^{l_i}$ and $\mathbf{v}_b$ are the normalized \textbf{[CLS]} embeddings of a text in language $i$ for $a$-th pairs and that of an image for the $b$-th pairs at each batch, which denotes the global representations of texts and images. In each iteration, we sample $N$ image-text pairs $(w^{l_i},v)$ from $D^{\mathrm{[VL]}}$ as one mini-batch. $\tau$ is a learnable temperature to scale the logits. 

We utilize Cross-lingual Contrastive Learning (CLCL)  on multilingual parallel text pairs to generalize cross-modal cross-lingual capabilities to a wide range of languages with language pivot. Similar to CMCL, CLCL is utilized to learn the global alignment of parallel bilingual text pairs with loss as
\begin{equation}
\mathcal{L}_{CLCL}=\mathbb{E}_{(w^{l_i},w^{l_i})\sim D^{[L]}}-\frac{1}{N}(\underbrace{ \sum_{a}^{N}\log\frac{{\exp{(\mathbf{w}_{a}^{l_i}}^\top}\mathbf{w}^{l_j}_{a}/\tau)}{\sum_{b}^{N}\exp{({\mathbf{w}_{a}^{l_i}}^{\top}\mathbf{w}^{l_j}_b/\tau)}}}_{\mathrm{language} \ i-\mathrm{to}-\mathrm{language} \ j}+\underbrace{\sum_{a}^{N}\log\frac{{\exp{({\mathbf{w}_{a}^{l_j}}^\top}}\mathbf{w}^{l_i}_{a}/\tau)}{\sum_{b}^{N}\exp{({{\mathbf{w}_a^{l_j}}^\top}\mathbf{w}^{l_i}_b/\tau)}}}_{\mathrm{language} \ j-\mathrm{to}-\mathrm{language} \ i}) 
\end{equation}
where $\mathbf{w}_{a}^{l_i}$ and $\mathbf{w}_{b}^{l_j}$ are the normalized \textbf{[CLS]} embedding of a text in the $a$-th and $b$-th bilingual pairs in language $i$ and language $j$. $w^{l_i}$ and $w^{l_j}$ are translated texts between two languages. 
\subsubsection{Cross-lingual Cross-modal fine-grained Understanding}
We employ two widely-used pre-training tasks in VLP (i.e., Image-Text Matching (ITM) and Masked Language Modeling (MLM) ) to further align vision and languages in a fine-grained semantic granularity. ITM attempts to predict whether the image-text pairs are matched or not. We use the outputs of a special token [MM\_CLS] in the cross-lingual cross-modal encoder as the joint representation of image-text pairs and feed it into the binary classifier with sigmoid function to predict a match score of image-text pairs $s_{\theta}(w^{l_i},v)$, $\theta$ is the learnable parameter of our framework. We use a binary cross-entropy loss as
\begin{equation}
\mathcal{L}_{ITM}(\theta)=\mathbb{E}_{(w^{l_i},v)\sim D^{[\mathrm{VL}]}}-[y\log{s_{\theta}(w^{l_i},v)}+(1-y)\log(1-s_{\theta}(w^{l_i},v))]
\end{equation}
where $y\in \{0,1\}$ denotes whether the image-text pairs is matched, and $w^{l_i}$ denotes texts in $i$ language. $(w^{l_i},v)$ consists of positive and negative pairs sampled from cross-lingual image-text pairs, and negative samples are constructed by replacing image or text in matched pairs. The loss function for MLM loss is 
\begin{equation}
\mathcal{L}_{MLM}(\theta)=\mathbb{E}_{(w^{l_{i}},v)\sim D^{[\mathrm{VL}]}}-\sum_{m \in M}\log{P_{\theta}(w^{l_{i}}_{m}|w^{l_{i}}_{ \textbackslash
 m},v)}
\end{equation}
where $w^{l_i}_{m}$ denotes the $m$-th masked word, and $M$ is the set of masked words for each text-image pair. $w^{l_{i}}_{\textbackslash
m}$ is texts with mask words.

\subsubsection{Cross-lingual Cross-modal Generation}
We use Prefix Language Modelling with and without the visual condition (vPLM and PLM) as generation pre-training tasks to learn cross-lingual cross-modal representation across monolingual texts and images. Specifically, Prefix Language Modeling (PLM) splits the multilingual texts of length $T$ into two parts (prefix sequence $w^{l_i} < T_{p}$ and suffix sequence $w^{l_i} \geq T_{p}$), then feeds them to the encoder and decoder, respectively. vPLM and PLM loss function is 

\begin{equation}
\mathcal{L}_{\mathrm{vPLM}}(\theta)=\mathbb{E}_{(w^{l_{i}},v)\sim D^{[\mathrm{VL}]}} -\sum_{t=T_{p}}^{T}\log{p_{\theta}(w^{l_{i}}_{t}|w_{[T_{p},t)}^{l_{i}},w_{t<T_{P}}^{l_i},v)}, \quad
\end{equation}
\begin{equation}
\mathcal{L}_{\mathrm{PLM}}(\theta)=\mathbb{E}_{w^{l_{i}}\sim D^{[\mathrm{L}]}} -\sum_{t=T_{p}}^{T}\log{p_{\theta}(w^{l_{i}}_{t}|w_{[T_{p},t)}^{l_{i}},w_{t<T_{P}}^{l_i})} 
\end{equation}

, where $T_{p}$ is length of prefix sequence. vPLM and PLM attempt to generate the suffix sequence of texts $w^{l_i}$ given by the prefix sequence with or without visual condition. Moreover, $T_p$ is sampled as 0 or the length of 20\% tokens of the whole sentence with equal probabilities. Therefore, vPLM acts as an image captioning task in a special case $T_p=0$.

Since vPLM learns cross-lingual cross-modal representation only with the visual pivot, we adopt Machine Translation (MT) and masked Multimodal Machine Translation (mMMT) to further enhance cross-lingual cross-modal learning leveraging bilingual text image-text pairs. Similarly, the mMMT is identical to machine translation with the visual condition, which aims to translate the source language with/without the images to the target language. Compared to conventional MMT, we employ mMMT by randomly replacing textual tokens in the source language with masked tokens (for such masked tokens, we also perform MLM on them) to enforce the interaction between images and texts in target language. For a similar purpose mentioned in Section \ref{Understanding} of extending to a wide range of languages, we follow standard Machine translation (MT) on the bilingual text pairs $(w^{l_{i}},w^{l_{j}})$ to enhance the cross-lingual alignment. The loss function for mMMT and MT is:

\begin{equation}
    \mathcal{L}_{\mathrm{mMMT}}(\theta)=\mathbb{E}_{(w^{l_{i}},w^{l_{j}},v)\sim D^{[\mathrm{VL}]}} -\sum_{t=0}^{T_{j}}\log{p_{\theta}(w ^{l_{j}}_{t}|w_{[0,t)}^{l_{j}},w^{l_i}_m,v)}, \quad
    \end{equation}
\begin{equation}
    \mathcal{L}_{\mathrm{MT}}(\theta)=\mathbb{E}_{(w^{l_{i}},w^{l_{j}})\sim D^{[\mathrm{L}]}} -\sum_{t=0}^{T_{j}}\log{p_{\theta}(w^{l_{j}}_{t}|w_{[0,t)}^{l_{j}},w^{l_i})}
\end{equation}

Where $T_j$ is the length of target texts, $w^{l_j}_{t}$ is the $t$-th token in target texts $w^{l_j}$, and $w_{m}^{l_i}$ is source texts with masked tokens.

\section{Experiment}
ERNIE-UniX$^{2}$ is jointly trained on text-only and text-image datasets. We evaluate ERNIE-UniX$^{2}$ across cross-lingual, cross-modal and cross-lingual cross-modal understanding and generation tasks. For understanding tasks,
 we evaluate on Multi30k \cite{elliott-etal-2016-multi30k} for cross-lingual cross-modal retrieval, XVNLI \cite{xvnli-lingual} and xGQA \cite{xGQA2021} from a new cross-lingual cross-modal benchmark IGLUE \cite{iglue2022} for complex cross-lingual vision-language tasks. For generation tasks, we evaluate on COCO \cite{mscoco,cococn} for image captioning in English and Chinese and Multi30k \cite{elliott-etal-2016-multi30k} for Multimodal Machine Translation (English-German). Moreover, we also evaluate our model on Tatoeba \cite{tatoeba} for mono-modal cross-lingual text retrieval.
 
\subsection{Pre-training Datasets}
\paragraph{Cross-lingual Image-text Datasets}
We utilize two cross-modal datasets, Conceptual Caption 3M (\textbf{CC3M}) and Conceptual 12M, (\textbf{CC12M}) \cite{cc12m} and one cross-lingual cross-modal dataset \textbf{WIT} \cite{wit2021} as our cross-lingual cross-modal dataset. We use translated caption of CC3M in UC$^2$ \cite{UC22021} and translate CC12M to five languages (German, French, Czech, Japanese, and Chinese). The Wikipedia-based Image-Text (WIT) \cite{wit2021} is mined from Wikipedia, including 37.6 million entity-rich image-text examples with 11.5 million unique images across 108 Wikipedia languages. We only use \textit{reference} of WIT as the corresponding caption in our training datasets, which contain 14M unique image-text pairs due to removing invalid URLs. The total of cross-lingual cross-modal datasets is 29M (CC3M, CC12M, WIT).
\paragraph{Multilingual Text-only Datasets}
To enhance learning in a wide range of languages, we use two text-only datasets CC100 and OPUS. CC100 corpus is a monolingual corpus \cite{wenzek-etal-2020-ccnet} including 111 languages and 5 romanized languages. OPUS is a bilingual corpus collecting from OPUS \cite{opus-2012}, including MultiUN \cite{ziemski-etal-2016-united}, IIT Bombay \cite{kunchukuttan-etal-2018-iit}, and WikiMatrix \cite{wikimatrix}. 

We train ERNIE-UniX$^2$ on these datasets using different pre-training tasks and list the details of each dataset used for pre-training tasks in Table \ref{pre_trainingtask_data}. 

\subsection{Downstream Tasks}
\paragraph{Cross-lingual Image-text Retrieval}
We evaluate ERNIE-UniX$^2$ on Multi30k \cite{elliott-etal-2016-multi30k,elliott-EtAl:2017:WMT,barrault2018findings} for cross-lingual image-text retrieval with zero-shot and fine-tuning settings. Multi30k \cite{elliott-etal-2016-multi30k} contains 31,783 images with four languages captions per image, extended from Flickr30K \cite{flicker30K} by manually translating the English captions to German, French and Czech.
\paragraph{Cross-lingual Visual Natural Language Entailment}
Recent cross-lingual cross-modal benchmark IGLUE \cite{iglue2022} proposes a new task of Cross-lingual Visual Natural Language Inference (XVNLI), aiming to predict the relationship between given images and texts across different languages. XVNLI combines the text-only dataset SNLI \cite{snli2015} with its cross-modal \cite{xvnli-modal} and cross-lingual \cite{xvnli-lingual} counterparts. Following the same zero-shot settings as IGLUE, we fine-tune ERNIE-UniX$^2$ on training splits of XVNLI with English-only image-text pairs and transfer it to the other four languages with zero-shot settings.
\paragraph{Cross-lingual Grounded Question Answering }
The Cross-lingual Grounded Question Answering (xGQA) \cite{xGQA2021} dataset extends the established English GQA \cite{GQA2019} dataset to 7 languages. We model xGQA as a multi-label classification task with 1,842 labels following IGLUE \cite{iglue2022} benchmark. Following the same zero-shot settings as IGLUE, we fine-tune ERNIE-UniX$^2$ on xGQA with English-only image-text pairs and transfer it to the other seven languages for zero-shot evaluation. 
\paragraph{Image Captioning}
The image captioning task attempts to generate natural language descriptions of input images. We consider two benchmarks named COCO-CN \cite{cococn} and MSCOCO \cite{mscoco} to evaluate ERNIE-UniX$^2$. For MSCOCO \cite{mscoco}, we keep the same train/test split as the Karpathy split \cite{Karpathy-split}, including 123,000 images, each annotated with five sentences. COCO-CN \cite{cococn} extends COCO captions to Chinese, which contains 20,342 images and 27,218 Chinese captions.
\paragraph{Multimodal Machine Translation}
Multimodal machine translation aims to translate image captions from the source language to the target language, where the images are considered as grounding signals. We consider a benchmark Multi30k \cite{elliott-etal-2016-multi30k}, which aims to translate English source sentences into its German human translation with the corresponding images.
\paragraph{Cross-lingual Text Retrieval}
We use a subset of the Tatoeba \cite{tatoeba} dataset to evaluate ERNIE-UniX$^2$ on cross-lingual text retrieval, consisting of 1,000 English-aligned sentence pairs covering 36 languages for cross-lingual Sentence Retrieval.
\subsection{Implementation Details}
\subsubsection{Pre-training Settings}
The cross-lingual cross-modal encoder-decoder of ERNIE-UniX$^{2}$ are initialized from mBART \cite{mbart}. For dual-encoder, we initialize the visual encoder with ViT-base \cite{vit2020} and the textual encoder with mBART. We train our model on 32 Nvidia A100 GPUs for 60K steps with a total batch size of 4096. We adopt Adam optimizer \cite{adamw} with a base learning rate of 10$^{-4}$, which is warmed up linearly in 4K steps and decayed to 10$^{-6}$ with a cosine schedule. The detailed hyperparameter of our pre-trained model is in Table \ref{hyperp_tabel} in the appendix.
\subsubsection{Evaluation Settings}
We evaluated  ERNIE-UniX$^{2}$ on six cross-lingual cross-modal tasks with zero-shot and fine-tuning settings for generation and understanding. For cross-lingual image-text retrieval and cross-lingual text retrieval, we only employ dual encoders of ERNIE-UniX$^{2}$ to evaluate our model both with zero-shot and fine-tuning settings. ERNIE-UniX$^{2}$ is fine-tuned with encoder-only architecture on other understanding tasks (XVNLI and xGQA) through appending an MLP layer on outputs of the first special token [MM\_CLS] to predict the target label. For generation tasks, image captioning and multimodal machine translation, we use the whole module of ERNIE-UniX$^{2}$ with fine-tuning settings and generate the text with 4 beam size. The details of fine-tuning settings in 
Table \ref{hyperp_tabel}.

\begin{table*}[]

\resizebox{\textwidth}{!}{

\begin{tabular}{@{}llccccccccclccccl@{}}
\toprule
\multicolumn{2}{c}{\multirow{4}{*}{Model}} & \multicolumn{2}{c}{NLI}         & \multicolumn{2}{c}{QA}          & \multicolumn{11}{c}{Retrieval}                                                                                                                                                                                                                                                                     \\ \cmidrule(l){3-4} 
\cmidrule(l){5-6}
\cmidrule(l){7-17}

\multicolumn{2}{c}{}                       & \multicolumn{2}{c}{XVNLI}       & \multicolumn{2}{c}{xGQA}        & Tatoeba        & \multicolumn{5}{c}{Multi30k zero-shot}                                                                                                                    & \multicolumn{5}{c}{Multi30k fine-tune}                                                                                \\
\multicolumn{2}{c}{}                       & EN             & 4-avg          & EN             & 7-avg          & 36-avg         & EN                        & DE                                 & FR                        & CS                                 & \multicolumn{1}{c}{Avg} & EN                        & DE                        & \multicolumn{1}{l}{FR}    & \multicolumn{1}{l}{CS}    & Avg   \\
\multicolumn{2}{c}{}                       & \multicolumn{2}{c}{Accuracy}         & \multicolumn{2}{c}{Accuracy}         & Accuracy            & \multicolumn{5}{c}{meanRecall}                                                                                                                                    & \multicolumn{5}{c}{meanRecall}                                                                                                \\ \midrule

(1)       & VECO                     & -              & -              & -              & -              & 86.91          & -                         & -                                  & -                         & -                                  & -                                  & -                         & -                         & -                         & -                         & \multicolumn{1}{c}{-}          \\
(2)       & ERNIE-M                        & -              & -              & -              & -              & 93.30          & -                         & -                                  & -                         & -                                  & -                                  & -                         & -                         & -                         & -                         & \multicolumn{1}{c}{-}          \\
(3)       & mUNITER                        & 76.38          & 53.69          & 54.68          & 9.97           & -              & -                         & -                                  & -                         & \textbf{-}                         & \textbf{-}                         & -                         & -                         & -                         & -                         & \multicolumn{1}{c}{\textbf{-}} \\
(4)       & xUNITER                        & 75.77          & 58.48          & 54.83          & 21.72          & -              & -                         & -                                  & -                         & -                                  & -                                  & -                         & -                         & -                         & -                         & \multicolumn{1}{c}{-}          \\
(5)       & M$^{3}$P                            & 76.38          & 58.25          & 55.19          & 28.17          & -              & 57.90                     & 36.80                              & 27.10                     & 20.40                              & \multicolumn{1}{l}{35.55}          & 87.40                     & 82.70                     & 73.90                     & 72.20                     & 79.05                          \\
(6)       & UC$^{2}$                            & 76.89          & 62.05          & 53.75          & 29.35          & -              & -                         & -                                  & -                         & -                                  & \multicolumn{1}{l}{}               & 88.20                     & 84.50                     & 83.90                     & 81.20                     & 84.45                          \\
(7)       & MURAL TrTrain(CC12m)+EOBT      & \textbf{-}     & \textbf{-}     & -              & -              & -              & \textbf{80.90}                     & \multicolumn{1}{l}{76.00}          & \multicolumn{1}{l}{\textbf{75.70}} & \multicolumn{1}{l}{68.20}          & \multicolumn{1}{l}{75.20}          & \multicolumn{1}{l}{\textbf{91.00}} & \multicolumn{1}{l}{\textbf{87.30}} & \multicolumn{1}{l}{\textbf{86.40}} & \multicolumn{1}{l}{82.40} & \textbf{86.77}                          \\
(8)       & ERNIE-UniX$^2$                          & \textbf{87.73} & \textbf{77.42} & \textbf{56.68} & \textbf{45.25} & \textbf{93.82} & \multicolumn{1}{l}{80.06} & \multicolumn{1}{l}{\textbf{76.60}} & \multicolumn{1}{l}{74.45} & \multicolumn{1}{l}{\textbf{72.27}} & \multicolumn{1}{l}{\textbf{75.85}} &

\multicolumn{1}{l}{88.80} & \multicolumn{1}{l}{84.71} & \multicolumn{1}{l}{84.05} & \multicolumn{1}{l}{\textbf{82.45}} & 85.00                          \\ \hline

\end{tabular}

}

\caption[caption]{ The results of cross-lingual cross-modal pre-training methods on understanding downstream tasks. EN means that the results are obtained by fine-tuning the pre-trained model on English-only datasets, for XVNLI, xGQA, Retrieval. X-avg means the zero-shot results for the average of X languages with an English-only fine-tuned model.  We use the average accuracy of 36 languages for Tatoeba. For image-text retrieval on Multi30k, meanRecall means the average of R@1, R@5, R@10 for image-to-text and text-to-image and avg means the average of meanRecall for four languages.} 

\label{table1}

\end{table*}

\begin{table*}[]
\centering
\resizebox{0.7\textwidth}{!}{
\begin{tabular}{@{}llccccccccc@{}}
\toprule
\multicolumn{2}{c}{\multirow{4}{*}{Model}} & \multicolumn{8}{c}{Image Captioning}                                               & Translation   \\ 

\cmidrule(l){3-10} 
\cmidrule(l){11-11} 

\multicolumn{2}{c}{}                       & \multicolumn{4}{c}{MSCOCO}                              & \multicolumn{4}{c}{COCO-CN}   & Multi30k      \\
\multicolumn{2}{c}{}                       & \multicolumn{4}{c}{En}                                & \multicolumn{4}{c}{Zh}     & En-De         \\
\multicolumn{2}{c}{}                       & B@4           & M             & C     & S             & B@4  & M    & R    & C     & B@4           \\ \midrule
(1)              & OSCAR                   & 36.5          & -             & 123.7 & -             & -    & -    & -    & -     & -             \\
(2)              & E2E-VLP                 & 36.2          & -             & 117.3 & -             & -    & -    & -    & -     & -             \\
(3)              & ERNIE-ViLG              & -             & -             & -     & -             & \textbf{50.0} & \textbf{31.6} & \textbf{60.3} & \textbf{138.2} & -             \\
(4)              & SimVLM                  & 40.6          & 33.7          & \textbf{143.3} & 25.4          & -    & -    & -    & -     & 47.6          \\
(5)              & ERNIE-UniX$^{2}$                   & \textbf{40.7} & \textbf{39.8} & 133.1 & \textbf{27.0} & 48.3 & 30.1 & 58.7 & 133.5 & \textbf{49.3} \\ \bottomrule
\end{tabular}

}

\caption[caption]{
Results on generation tasks of ERNIE-UniX$^{2}$ compared to other single language cross-modal pre-training models. 
We use BLEU@4 for Multi30k on multimodal machine translation. 
We use BLEU@4, METEOR, CIDEr and SPICE on MSCOCO and COCO-CN for image captioning. (B@4: BLEU@4, M: METEOR, C: CIDEr, S: SPICE)} 

\end{table*}

\subsection{Overall Results}
We evaluate ERNIE-UniX$^{2}$ on several understanding and generation tasks compared to previous state-of-the-art (SOTA) methods.  

For the understanding task, as shown in Table \ref{table1},  ERNIE-UniX$^{2}$ achieves new state-of-the-art results, outperforming VECO  \cite{luo2021veco}, ERNIE-M \cite{erniem2020}, mUNITER  \cite{uniter2021}, xUNITER \cite{uniter2021}, M$^3$P \cite{m3p2020}, UC$^2$ \cite{UC22021} and MURAL \cite{mural2021}. Specifically, for zero-shot cross-lingual transfer, ERNIE-UniX$^{2}$ substantially outperforms UC$^2$ on XVNLI and xGQA, with an average improvement of over 15.37\% and 11.28\%. For the cross-lingual text retrieval task, ERNIE-UniX$^{2}$ achieves a score of 93.82 in the Tatoeba dataset, outperforming the previous SOTA result of ERNIE-M. As well as, we use two different settings of zero-shot and all-language fine-tuning to compare ERNIE-UniX$^{2}$ with M$^3$P, UC$^2$ and MURAL on Multi30k image-text retrieval. ERNIE-UniX$^{2}$ also slightly outperforms MURAL, the prior state-of-the-art in the zero-shot by 0.65\% averaged recall across four languages. This demonstrates that our unified pre-training approach is competitive with understanding-based models. 

For the generation task, as the results are shown in Table \ref{table1}, 
ERNIE-UniX$^{2}$ outperforms SimVLM \cite{simvlm2021} on 3 out of 4 metrics of the image captioning task on MS-CoCo. Our model also achieves comparable results to the previous SOTA model ERNIE-ViLG \cite{ernievilg2021} for image captioning on COCO-CN. We note that ERNIE-ViLG is a 10 billion parameter model pre-trained on hundreds of millions of image-text pairs. In the image translation of Multi30k from English to German, ERNIE-UniX$^{2}$ has a noticeable improvement compared to SimVLM by 1.70\%, demonstrating that large and diverse translation pairs can learn better cross-lingual cross-modal encoder-decoder. These experiments demonstrate that our model can achieve superior performance through the unified cross-lingual cross-modal encoder-decoder architecture.

\begin{table*}[]

\resizebox{0.95\textwidth}{!}{
\centering
\begin{tabular}{@{}ll|ccccccc@{}}
\toprule
\multicolumn{2}{c|}{}                          & \multicolumn{2}{c}{xGQA} & Multi30k retrieval & \multicolumn{2}{c}{MSCOCO} & Multi30k translation         &                       \\ 
\cmidrule(lr){3-4}
\cmidrule(lr){5-5}
\cmidrule(lr){6-7}
\cmidrule(lr){8-8}
\multicolumn{2}{c|}{\multirow{-2}{*}{Methods}} & EN        & 7-avg    & meanRecall                 & B@4          & C            & B@4                          & \multirow{-2}{*}{Avg} \\ \midrule
(0)                 & ERNIE-UniX$^2$                     & 55.08     & 40.44        & 81.45              & 40.20        & 131.30       & 49.30                        & 66.30                 \\ \midrule
(1)                 &  \quad w/o ITM                 & 53.61     & 40.53        & 81.39              & 40.20        & 131.60       & 49.50                        & 66.14                 \\
(2)                 &  \quad w/o MLM                 & 53.02     & 39.94        & 81.76              & 39.20        & 130.00       & 49.30 & 65.54                 \\
(3)                 &  \quad w/o CMCL                & 53.45     & 38.73        & 68.94              & 38.90        & 127.80       & 49.00                        & 62.80                 \\ \midrule
(4)                 &  \quad w/o mMMT                & 54.51     & 40.99        & 81.07              & 40.20        & 131.40       & 49.10                        & 66.21                 \\
(5)                 &  \quad w/o vPLM               & 52.09     & 39.30        & 81.63              & 39.60        & 128.80       & 48.90                        & 65.05                 \\ \bottomrule
\end{tabular}

}
\caption[caption]{ 
The Ablation study of proposed pre-training tasks in ERNIE-UniX$^2$. "ITM" refers to Image-Text Matching. "MLM" refers to Masked Language Modeling. "CMCL" refers to Cross-model Contrastive Learning. "mMMT" refers to masked Multimodal Machine Translation. "vPLM" refers to visual Prefix Language Modeling. }.  

\label{tabel4}

\end{table*}

\subsection{Ablation Studies}
\paragraph{The Effect of Pre-training Tasks}
To validate the effectiveness of pre-training tasks in ERNIE-UniX$^{2}$, we conduct ablation studies and report results in Table \ref{tabel4}. Concretely, we evaluate our ERNIE-UniX$^{2}$ on two understanding tasks (i.e.,  cross-lingual image-text retrieval on Multi30k and xGQA) and two generation tasks (i.e., image captioning on MSCOCO and multimodal machine translation on Multi30k). For xGQA, following the same zero-shot settings as IGLUE, we present English-only fine-tuned results and transfer the model to other 7 languages to obtain zero-shot results. We present the all-language fine-tuned results on Multi30k for cross-lingual image-text retrieval across 7 languages. For generation tasks, we present fine-tuned results on image captioning on MSCOCO following Karpathy splits and multimodal machine translation on Multi30k. All compared models are pre-trained for 50K steps with a batch size of 512.

As shown in Table \ref{tabel4}, ITM presents a significant improvement on English-only fine-tuning on xGQA yet a slight improvement across all tasks, which illustrates the importance of the fine-grained cross-modal alignment. Then, we analyze the effect of (2) MLM and (3) CMCL. The use of the CMCL objective significantly improved the performance of the Multi30k retrieval task. It is further proved that understanding and generation pre-training tasks are mutually reinforcing. Moreover, (4) mMMT improves multimodal machine translation and image-text retrieval. In addition, we study the contributions of (5) vPLM, and find it critical for all downstream tasks performance. These phenomenons indicate that the generation pre-training task improves not only the performance of the downstream generation task but also the performance of the downstream understanding task. 

\paragraph{The Effect of ERNIE-UniX$^2$ Framework}
For a fair comparison with the previous best methods (M$^3$P, UC$^2$), we reduce the parameters of ERNIE-UniX$^2$ to the same size using fewer Transformer layers and the same pre-training datasets. As M$^3$P and UC$^2$ also initialize their 12-layer transformer model from XLM-R \cite{xlmr2019}, we follow a similar initialization with the multilingual text encoder and cross-lingual cross-modal encoder initialized with the first and second 6 layers from XLM-R. As shown in Table \ref{tabel4}, ERNIE-UniX$^2\dagger$ performs better than M$^3$P and UC$^2$ in both the Multi30k retrieval task and xGQA task, demonstrating the superiority of our proposed framework for cross-lingual cross-modal pre-training.

\begin{table}[]
\centering

\centering

\begin{tabular}{@{}l|cc@{}}
\toprule
Modal & xGQA           & Multi30k retrieval \\ \midrule
M$^{3}$P   & 28.17          & 35.55              \\
UC$^{2}$   & 29.35          & -              \\
ERNIE-UniX$^{2}$ & \textbf{32.20} & \textbf{55.10}     \\ \bottomrule
\end{tabular}
\caption{The zero-shot results on xGQA and Multi30K retrieval  of ERNIE-UniX$^{2}$ compared to previous cross-lingual cross-modal models (M$^3$P, UC$^2$),  under settings of comparable parameters size. For xGQA, we use the average accuracy of 7 languages with an English-only fine-tuned model. For Multi30k image-text, we use mR (meanRecall) that is the average meanRecall of four languages. $\dagger$ is initialized by XLM-R} 

\label{tabel3}

\end{table}

\section{Conclusion}
We present a unified framework ERNIE-UniX$^2$  across languages, modalities, and tasks. ERNIE-UniX$^2$ integrates multiple pre-training paradigms with encoder-decoder architecture, which can seamlessly be fine-tuned on various generation and understanding downstream tasks. ERNIE-UniX$^2$  achieves SOTA results on several cross-lingual cross-modal generation and understanding tasks and competitive results for monolingual cross-modal generation in well-resourced language, which shows the effectiveness on varieties of downstream tasks. 
For the future work, we will further improve our unified framework through integrating more modalities and more pre-training paradigms. Also we will apply our model to single-modal tasks such as image classification for the visual modality. 
\bibliographystyle{unsrt}  
\bibliography{references}  

\clearpage

\appendix

\section{Hyperparameters for Pre-training and Fine-tuning}
Table \ref{hyperp_tabel} lists the hyperparameters for pre-training and the fine-tuning on XVNLI, xGQA, MSCOCO, COCO-CN, Multi30k retrieval and translation. 
\section{The datasets for Pre-training tasks}
We list the pre-training tasks used for each dataset in Table \ref{pre_trainingtask_data}.
\section{The results on XVNLI, xGQA and Tatoeba for each language}
We report the performance of the zero-shot on XVNLI and xGQA in Tabel \ref{tabel7} and Tatoeba in Table \ref{table9} for each language.

\begin{table}[]
\centering
\resizebox{1\linewidth}{!}{
\begin{tabular}{@{}l|ccccccc@{}}
\toprule
\multicolumn{1}{c|}{\multirow{2}{*}{\textbf{Hyperparameter}}} & Pre-training & \multicolumn{6}{c}{Fine-tuning} \\ \cmidrule(l){2-2} 
\cmidrule(l){3-8} 
\multicolumn{1}{c|}{} & - & xGQA & XVNLI & \begin{tabular}[c]{@{}l@{}}Multi30k\\  retrieval\end{tabular} & \begin{tabular}[c]{@{}l@{}}Multi30k \\ translation\end{tabular} & MSCOCO & COCO-CN \\ \midrule
batch size & 4096 & 1024 & 1024 & 4096 & 256 & 256 & 256 \\
learning rate & 1e-4 & \multicolumn{1}{c}{5e-5} & 5e-5 & 5e-5 & 5e-5 & 5e-5 & 5e-5 \\
warmup updates & 4000 & 2000 & 2000 & 50 & - & - & - \\
learning rate schedule & \multicolumn{4}{c}{Cosine Schedule Decaying to Zero} & \multicolumn{3}{c}{-} \\
AdamW weight decay & \multicolumn{7}{c}{0.1} \\
AdamW $\beta_{1}$ & \multicolumn{7}{c}{0.9} \\
AdamW $\beta_{2}$ & \multicolumn{7}{c}{0.999} \\ \bottomrule
\end{tabular}
}
\caption[caption]{A summary of pre-training and fine-tuning hyperparameters in ERNIE-UniX$^{2}$.}
\label{hyperp_tabel}
\end{table}
\begin{table}[]
\centering
\resizebox{0.7\linewidth}{!}{
\begin{tabular}{@{}ll|lccc@{}}
\toprule
Type &  & Pre-training Task & \#Image & \#Sample & \#Lang \\ \midrule
 & CC3M &  & 3M & 15M & 5 \\ \cmidrule(lr){2-2} \cmidrule(l){4-6} 
Image-texts & CC12M & CMCL, ITM, MLM, vPLM, mMMT & 12M & 60M & 5 \\ \cmidrule(lr){2-2} \cmidrule(l){4-6} 
 & WIT &  & 14M & 14M & 108 \\ \midrule
Texts & OPUS & CLCL, MT & - & 55M & 99 \\ \cmidrule(l){2-6} 
 & CC100 & PLM, MLM & - & 1.5TB* & 96 \\ \bottomrule
\end{tabular}
}
\caption{Statistics on the datasets of pre-training tasks. \#Image,\#Sample and \#lang denote the number of unique images, samples and languages. *For language data, we report its storage following \cite{erniem2020}}
\label{pre_trainingtask_data}
\end{table}

\begin{table}[H]
\centering
\resizebox{1\linewidth}{!}{
\begin{tabular}{@{}llllllllllll@{}}
\toprule
Models & \multicolumn{4}{c}{XVNLI} & \multicolumn{7}{c}{xGQA} \\ \cmidrule(l){2-5} \cmidrule(l){6-12} 
 & ARB & SPA & FRA & RUS & BEN & DEU & IND & KOR & POR & RUS & CMN \\ \midrule
mUNITER & 46.73 & 56.96 & 59.36 & 51.72 & 3.9 & 26.3 & 13.0 & 5.5 & 17.4 & 9.3 & 8.4 \\
xUNITER & 51.98 & 58.94 & 63.32 & 59.71 & 9.7 & 35.7 & 36.3 & 14.6 & 24.7 & 20.8 & 19.2 \\
M$^3$P & 56.19 & 57.47 & 69.67 & 64.86 & 19.2 & 35.6 & 29.6 & 28.5 & 35.7 & 32.9 & 31.1 \\
UC$^2$ & 55.24 & 58.85 & 56.36 & 62.54 & 20.8 & 45.1 & 28.1 & 24.3 & 27.6 & 31.9 & 33.9 \\
ERNIE-UniX$^2$ & \textbf{74.08} & \textbf{78.86} & \textbf{81.04} & \textbf{75.73} & \textbf{25.9} & \textbf{49.7} & \textbf{41.5} & \textbf{37.0} & \textbf{43.6} & \textbf{40.4} & \textbf{46.4} \\ \bottomrule
\end{tabular}
}
\caption{The zero-shot performance on xGQA and XVNLI with Accuracy. (i.e. English-only fine-tuning).}
\label{tabel7}
\end{table}

\begin{table}[H]
\centering
\resizebox{1\linewidth}{!}{
\begin{tabular}{l|cccccccccccccccccc}
\toprule
\textbf{Model}& \textbf{af}& \textbf{ar}& \textbf{bg}& \textbf{bn}& \textbf{de}& \textbf{el}& \textbf{es}& \textbf{et}& \textbf{eu}& \textbf{fa}& \textbf{fi}& \textbf{fr}& \textbf{he}& \textbf{hi}& \textbf{hu}& \textbf{id}& \textbf{it}& \textbf{ja} \\
\midrule
VECO & 80.9& 85.1& 91.3& 78.1& 98.5& 89.5& 97.4 & 94.8& 79.8& 93.1 & 95.4 & 93.7& 85.8& 94.2& 93.8 & 93.0& 92.2& 92.8 \\

\textsc{Ernie-M} & 92.6 & \textbf{94.3} & \textbf{96.6} & 89.2 & \textbf{99.7} & \textbf{96.8} & 98.8 & 92.5 & 87.4 & \textbf{96.0} & 97.1 & \textbf{96.5} & 90.1 & \textbf{97.9} & 95.5 & 95.7 & \textbf{95.2} & 96.9 \\

\textsc{ERNIE-UniX$^2$} & \textbf{97.2} & 91.8 & 95.3 & \textbf{90.5} & \textbf{99.7} & 96.5 & \textbf{98.9} & \textbf{98.0} & \textbf{95.6} & 95.7 & \textbf{97.8} & 96.2 & \textbf{92.2} & 97.3 & \textbf{96.8} & \textbf{96.3} & \textbf{95.2} & \textbf{97.4} \\

\midrule
\textbf{Model}& \textbf{jv}& \textbf{ka}& \textbf{kk}& \textbf{ko}& \textbf{ml}& \textbf{mr}& \textbf{nl}& \textbf{pt}& \textbf{ru}& \textbf{sw}& \textbf{ta}& \textbf{te}& \textbf{th}& \textbf{tl}& \textbf{tr}& \textbf{ur}& \textbf{vi}& \textbf{zh}  \\
\midrule
VECO & 35.1& 83.0& 74.1& 88.7& 94.8& 82.5& 95.9& 94.6& 92.2& 69.7& 82.4& 91.0& 94.7& 73.0& 95.2& 63.8& 95.1& 93.9 \\

\textsc{Ernie-M} & \textbf{65.2} & \textbf{94.9} & \textbf{88.0} & \textbf{94.1} & \textbf{98.5} & 90.8 & \textbf{98.1} & 94.5 & \textbf{95.7} & 68.4 & \textbf{91.8} & \textbf{97.9} & \textbf{98.4} & 86.0 & \textbf{98.3} & \textbf{94.9} & 98.1 & \textbf{96.7} \\

\textsc{ERNIE-UniX$^2$} & 60.0 & 92.1 & 85.9 & 93.0 & 98.4 & \textbf{92.5} & 97.4 & \textbf{95.6} & 95.1 & \textbf{79.0} & 90.2 & 96.2 & 97.5 & \textbf{90.5} & \textbf{98.3} & 92.9 & \textbf{98.2} & \textbf{96.7} \\

\bottomrule
\end{tabular}
}
\caption{The cross-lingual retrieval performance on Tatoeba results for each language.}
\label{table9}
\end{table}

\end{document}